# Implementation of Deep Neural Networks to Classify EEG Signals using Gramian Angular Summation Field for Epilepsy Diagnosis

K. Palani Thanaraj[1], B. Parvathavarthini[2], U. John Tanik[3], V. Rajinikanth[1,*], Seifedine Kadry[4], K. Kamalanand[5]

[1]Department of Electronics and Instrumentation Engineering, St. Joseph's College of Engineering, Chennai, India.
[2]Department of Computer Science and Engineering, St. Joseph's College of Engineering, Chennai, India.
[3]Department of Computer Science and Information Systems, Texas A&M University-Commerce, USA.
[4]Department of Mathematics and Computer Science, Beirut Arab University, Lebanon.
[5]Department of Instrumentation Engineering, MIT Campus, Anna University, Chennai, India.

*Corresponding author email address: v.rajinikanth@ieee.org

**Abstract:** This paper evaluates the approach of imaging time-series data such as EEG in the diagnosis of epilepsy through Deep Neural Network (DNN). EEG signal is transformed into an RGB image using Gramian Angular Summation Field (GASF). Many such EEG epochs are transformed into GASF images for the normal and focal EEG signals. Then, some of the widely used Deep Neural Networks for image classification problems are used here to detect the focal GASF images. Three pre-trained DNN such as the AlexNet, VGG16, and VGG19 are validated for epilepsy detection based on the transfer learning approach. Furthermore, the textural features are extracted from GASF images, and prominent features are selected for a multilayer Artificial Neural Network (ANN) classifier. Lastly, a Custom Convolutional Neural Network (CNN) with three CNN layers, Batch Normalization, Max-pooling layer, and Dense layers, is proposed for epilepsy diagnosis from GASF images. The results of this paper show that the Custom CNN model was able to discriminate against the focal and normal GASF images with an average peak Precision of 0.885, Recall of 0.92, and F1-score of 0.90. Moreover, the Area Under the Curve (AUC) value of the Receiver Operating Characteristic (ROC) curve is 0.92 for the Custom CNN model. This paper suggests that Deep Learning methods widely used in image classification problems can be an alternative approach for epilepsy detection from EEG signals through GASF images.

**Keywords:** Electroencephalogram; Gramian Angular Summation Field; Deep Learning; Deep Neural Network; Convolutional Neural Network; Imaging Time Series data

## 1. Introduction

In the human body, the brain is a profoundly vital organ, responsible for the comprehensive monitoring and autonomous control of metabolic operations. Brain abnormalities, such as epilepsy, ischemic strokes, and brain tumors, may affect normal biological functions [1]. A suitable signal or imaging modality is recommended by the doctor if a brain abnormality is identified during screening operation. Diagnostic tools such as Electroencephalography (EEG), Magneto-encephalography (MEG), Computed Tomography (CT), Magnetic Resonance Imaging (MRI), and Positron Emission Tomography (PET) are some of the common modalities often used in the diagnosis of brain disorders [2-7]. However, EEG is widely used in preliminary analysis of brain functions owing to its cost-effectiveness and high temporal resolution. It records the electric potentials of neuron activations from different regions of the brain. Epilepsy is an abnormal condition of the brain, which leads to uncontrolled discharge of neuronal impulses from a specific brain region. It often leads to muscular convulsions known as epileptic seizures. Epileptologists are required to analyze the EEG recordings of the affected individuals to determine the severity of the seizures. Often this is a cumbersome task and leads to subjective inferences. To overcome this problem, a computer-based epilepsy diagnosis system is widely researched [8].

Recent literature suggests an automated epilepsy detection system that extracts temporal or frequency features from EEG and classifies it as either normal or focal (epileptic) EEG [9-12]. Machine Learning (ML) techniques such as Decision-Tree, Linear-Discriminant-Analysis (LDA), Support-Vector-Machine (SVM), k-NN, and Random Forest are widely used in epilepsy detection [13-15]. SVM classifier based on Radial-Basis-Function (RBF) has proven to be an effective binary classifier in epilepsy diagnosis from the EEG signal [16-18]. However, the accuracy of the classifier depends on the nature of the feature used for classifying EEG. Many earlier works have suggested a feature selection procedure to select the appropriate features for improving the classification performance of the classifier [19-22]. Thus, conventional methods of analyzing the EEG signal involve a considerable amount of manual supervision in selecting an appropriate feature for epilepsy detection [19, 20]. More recently, in the field of image classification, an automated feature extraction procedure is widely used using Convolutional-Neural-Network (CNN), and high classification accuracy has been reported in the literature. CNN-based Deep Learning (DL) procedures provide an effective generalization property in classifying images. Many such networks with simplified architecture are recently invented to solve a variety of image classification problems, and some of these architectures include the AlexNet [23], VGG16, and VGG19 [24]. The main advantage of these

DNN is that its capacity to re-train on any new dataset for the classification task. Moreover, these networks and can be run on specialized hardware for faster training on large datasets. The impact of CNN in image classification applications has motivated us to implement DL in EEG signal classification.

More recently, attempts to classify time-series signal using CNN has been tried in the works of Wang and Oates [25] and Hatami et al. [26]. In their work, the time-series signal is divided into smaller time epochs. Then it is converted to an image signal using mathematical transforms such as Recurrence Plots (RP), Markov Transition Field (MTF), Gramian Angular Summation Field (GASF), and Gramian Angular Difference Field (GADF).

In this research work, the application of DL in the classification of epileptic EEG using state-of-the-art pre-trained architectures, namely the AlexNet, Visual Geometry Group's VGG16, and VGG19 are studied using the benchmark EEG signals available in [27]. A custom CNN architecture is also proposed here for the classification of EEG. Later, the performance of the pre-trained CNN and the custom CNN is compared with the recent feature-based Deep-ANN proposed by Bakiya et al. [28,29].

The workflow of this research paper involves the following steps. First, the EEG dataset is divided into smaller time epochs (256 time samples), and for each EEG time epoch, GASF is implemented to convert the time-series signal into an RGB image. This is an image signal which is a representation of the time-series data. Many such GASF images are obtained for both normal and focal EEG epochs. Then the considered DNN architectures are used for classifying the normal and focal GASF images. In this work, a transfer learning approach is implemented in the case of pre-trained CNN. The pre-trained CNN architectures retain the weight configuration of the deeper neural layers, and the top layers are retrained for the GASF image classification problem.

The experimental work for the considered DNN architectures is implemented using the deep learning libraries available in Matlab and Python software environment. Performance metrics such as Precision, Recall, and F1-score, are calculated for all the networks for performance validation in epilepsy detection. Moreover, the Receiver Operator Characteristic (ROC) curve is also plotted to assess the classification performance of the DNN architectures studied in this paper.

## 2. Context

Recently, a substantial amount of research work has been done to examine various classes of images using deep learning techniques based on CNN. However, implementation of the Deep-Learning procedure for the bio-signals are not extensively researched compared to the medical images. Methods based on Recurrent Neural Network (RNN) such as Long-Short Term Memory (LSTM) and Gated Recurrent Unit (GRU) have been used in bio-signal processing. However, these DL methods are often difficult to train if the training data has a high noise component. Any change in signal amplitude due to noise can drastically affect the reliability of the network [30]. Moreover, the training time of RNN based signal detection system is more as the network has to learn long temporal dependencies from bio-signals such as EEG [31]. Thus, the information extraction from the signal is quite difficult compared to images. Hence, in recent years, the conversion of the existing medical signals into images using a class of procedures is widely discussed by the researchers. Furthermore, when the existing signal is transformed in the form of a picture (RGB/grayscale), an existing image classification procedure can be easily implemented using a pre-trained or customized CNN architectures [32-33].

Due to the availability of modern computing methodologies, DL based on DNN is widely adopted by the researchers to examine the medical images and signals. The work of Amin et al. implemented a deep CNN methodology to assess the brain MRI and attained better classification accuracy with the benchmark brain tumor dataset [4]. Abd-Ellah et al. proposed a five-layered region-based CNN architecture to classify the BraTS2013 database, validating the performance of the proposed CNN with the existing AlexNet, VGG16, and VGG19 which confirmed the proposed CNN offers better classification accuracy (99.55%) during the brain tumor analysis [34]. The recent work of Wang et al. implemented a customized CNN architecture to classify the mice microscopic images into normal, granuloma-fibrosis1, and granuloma-fibrosis2 and achieved better classification accuracies [35]. Gao et al. implemented a CNN structure to classify the CT brain pictures and achieved a classification accuracy of >84%. Further, the customized CNN architectures are also implemented to classify the medical signals, similar to the medical pictures [36]. Tripathy and Acharya implemented a DNN configuration to classify the EEG signals based on the RR-time series features [33]. The research work by Acharya et al. employs a deep CNN to recognize myocardial infarction using the patient's ECG signals. The recent research work by Acharya et al. proposed a Deep CNN to automate seizure detection based on EEG signals [37].

From the above works, it can be noted that the implementation of the transfer-learning and customized CNN architectures for the examination of medical images/signals have been reported with good classification accuracy. Hence, in recent years, examining signals in the form of images are widely executed by the researchers. Some of the works that involve transforming signals to images for classification include the application of S-transform for analyzing EMG [29], implementing the time-frequency (T-F) spectrum for the EEG [38,39] and the ECG signal classification[41-42]. After the signal to image conversion, these images can be examined using the CNN or any other DNN architecture trained to assess the images.

In the proposed research work, the benchmark EEG signal (normal and focal) [27] is initially converted into RGB images using the GASF technique. Initially, the most famous pre-trained CNN architectures, such as AlexNet, VGG16, and VGG19 are adopted to examine the considered dataset using the transfer learning approach. Then a custom CNN architecture is designed for EEG signal classification. Finally, a Deep-ANN architecture proposed by Bakiya et al. [29] is adopted, and the performances of all DNN models are compared.

3. EEG database

In this work, the benchmark Bern-Barcelona (Bern) database available in [27] is considered for the examination. This database is initially prepared by Andrzejak et al. from the normal and epilepsy volunteers. Based on the amplitude level of the EEG, it can be categorized as normal and epilepsy cases. The examination of the entire EEG samples existing in the dataset is quite complex and time-consuming. Hence, to minimize the analysis time/cost, a data-splitting technique is implemented to partition the test EEG signal into a time series segments of 256 samples, which can then be converted into 256x256 pixel-sized images.

4. Gramian Angular Summation Field (GASF)

Wang and Oates [25] proposed GASF to transform time-series signals into images. The encoding process involves normalizing the input time series data into the range of [-1,1]. The normalized or scaled time-series signal is then converted to a polar co-ordinate from Cartesian co-ordinate. This transform retains the temporal information of the input signal. The signal is warped in the transform domain. After this, each time point in polar co-ordinates is compared with every other point for temporal correlation. This is done by using trigonometric cosine function, which leads to the Gramian matrix of dimension [n,n], where n is the number of sample points of the EEG time epoch.

Let, $T = \{t_1, t_2, ..., t_n\}$ denotes a signal with n-samples and the $T$ can be rescaled to have the interval [-1,1] which can be given by the form below;

$$\tilde{T}_0^i = \frac{t_i - min(T)}{max(T) - min(T)} \qquad (1)$$

Then the angle φ is computed using the equation,
$$\varphi = \arccos(\tilde{T}_o^i) \qquad (2)$$

The temporal correlations of the adjacent points (i,j) are computed by determining the summation of the angle, thus leading to the Gram matrix called Gramian Angular Summation Field. This can be written as,

$$GASF = \left[ cos\left(\varphi_i + \varphi_j\right) \right] \qquad (3)$$

This technique can be adapted to transform a chosen time-series sample into an image. Other details on GASF can be found in [25]

In this work, 256 samples are taken for each EEG epoch, thus leading to a GASF image of [256,256]. As the sum of the angle of the data points is considered, this leads to the Gramian Angular Summation Field. The Bern EEG dataset consists of a single-channel recording of normal and focal EEG signals. The EEG signal is split by dividing the entire range (10240 samples/sampling rate=512Hz) into smaller time segments of 256 samples for the computation of GASF. A total count of 312 normal and 312 focal GASF is obtained from EEG data files for training the deep-learning network. In addition, 78 normal and 78 focal GASF are obtained for validation and performance analysis of the CNN architectures. In summary, a total of 780 images are collected from various EEG recordings of the Bern-EEG dataset, which consists of 199,680 time samples. Moreover, data augmentation based on different image transformations such as image rotation, shifting, and shearing is implemented for data generation for effective training of the DNN models.

5. Methodology

The Deep Neural Networks implemented in this research work to detect epilepsy based on EEG are presented in this section. Initially, a transfer learning-based CNN is implemented using AlexNet, VGG16, and VGG19, and the adopted block diagram for this task is depicted in Fig 1. Later, a custom CNN model is proposed for EEG signal classification, which is shown in Fig. 2. Thirdly, a feature-based Deep-ANN is presented for epilepsy detection which is shown in Fig. 4

The workflow of the paper starts with the preprocessing stage. Here, the EEG segments are split into smaller epochs of 256 samples. Then the EEG epochs are transformed into GASF images, which are given a pseudocolor to obtain an RGB color image.

5.1 Transfer Learning-based CNN

In the literature, pre-trained CNN architectures are readily available for the purpose of implementing the image classification task. These architectures are trained on large image datasets such as ImageNet for classification of everyday objects [40]. Each CNN architecture consists of a different number of processing layers, such as the convolutional layers, MaxPooling layers, and Dense Layers. In this work, the CNN architectures, such as the AlexNet [23], VGG16, and VGG19 [24] are adopted for the EEG classification task. The pre-trained models support the transfer-learning technique. Here the lower layers which are trained on ImageNet are retained, and only the top layers of the network are retrained for the GASF based epilepsy detection. Moreover, transfer learning of the pre-trained models also involves less training time as only the high-level features of the network in the top layers are trained for the GASF image classification.

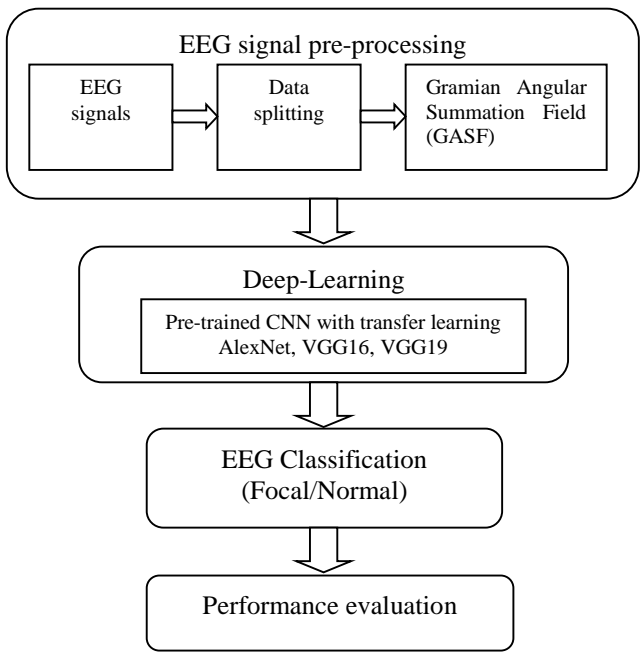

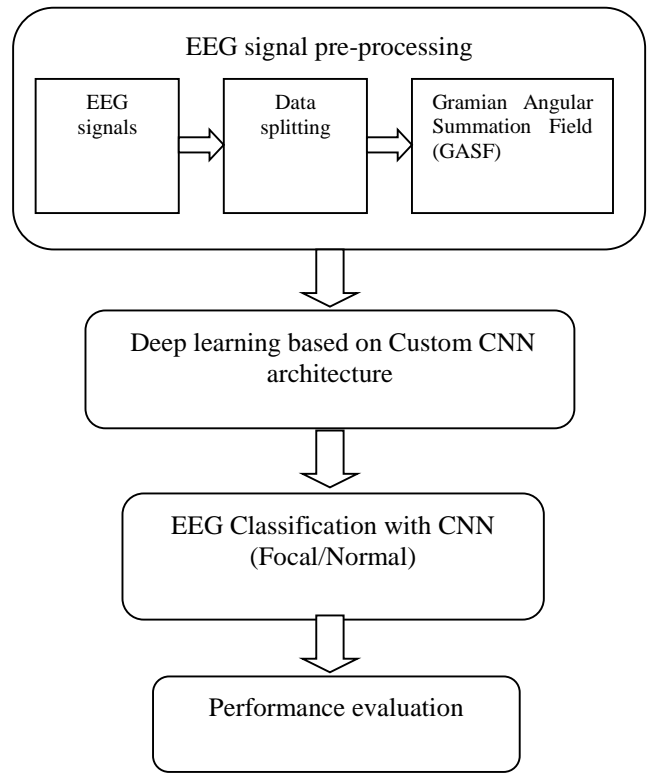

**Fig.1. Block diagram of Pre-trained CNN based EEG assessment**

**Fig.2. Block diagram of Custom CNN based EEG assessment**

### 5.2 Custom CNN for Epilepsy Detection

A multilayer deep neural network based on the convolutional neural net is proposed here for epilepsy detection from GASF images. In this work, we propose three convolutional layers with the ReLU activation function, followed by Batch Normalization (BN) and the Max pooling layer. After this, high-level features are extracted by the Dense layers with Sigmoid activation function. The final classification layer consists of a SoftMax activation function for the binary classification of normal and focal GASF. Table 1 provides the detail configuration of the various processing blocks of the proposed CNN model.

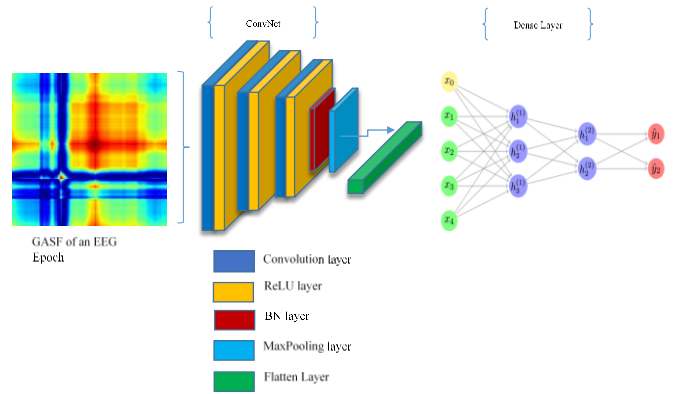

**Fig.3. Illustration of different deep learning blocks of the custom CNN architecture**

Fig.3. provides an illustration of the proposed CNN based custom epilepsy detection model, which takes GASF images of normal and focal EEG signals and classifies them.

The training of the custom CNN model involves data augmentation. Here the input data samples are augmented by image transformations for more data samples for better training of the network.

**Table 1 Configuration of the Custom CNN architecture for EEG analysis**

| Layer Name Kernel Size, # Filters | Input Size | Output Size | Stride | Activation Function |
|---|---|---|---|---|
| Input Image | [224,224,3] | [224,224,3] | - | - |
| ConvNet-1 [3x3], 32 | [224,224,3] | [222,222,32] | 1 | ReLU |
| ConvNet-2 [3x3], 64 | [222,222,32] | [110,110,64] | 2 | ReLU |
| ConvNet-3 [3x3], 64 | [110,110,64] | [54,54,64] | 2 | ReLU |
| Batch Norm | [54,54,64] | [54,54,64] | - | - |
| Max Pooling [2x2] | [54,54,64] | [27,27,64] | 1 | - |
| Flatten | [27,27,64] | [46656,1] | - | - |
| Dense-1 | [46656,1] | [1024,1] | - | Sigmoid |
| Dense-2 | [1024,1] | [512,1] | - | Sigmoid |
| Dense-3 | [512,1] | [2,1] | - | SoftMax |
| Classifying Layer | [Normal, Focal] | | | |

Moreover, the training sequence of the custom model involves checkpoints. Here the validation accuracy is monitored for 10 trials. The weight parameters for which the validation accuracy is highest is stored for the performance assessment of the custom CNN model.

### 5.3 Feature Based Deep ANN for Epilepsy Detection

The third DNN model attempted in this work for EEG analysis in epilepsy detection is adopted from the works of Bakiya et al. [29]. Their work involved the extraction of various textural features from the Time-Frequency spectrum of EMG signals. It involved using various signal transform techniques such as S-Transform (ST), Wigner-Ville Transform (WVT), Short-time Fourier Transform (STFT), and Synchro-Extracting Transform (SET). Later they employed a multilayer Deep ANN based classifier for the detection of Amyotrophic Lateral Sclerosis (ALS). In this research work, the textural features are obtained from the GASF images of normal and focal EEG signals.

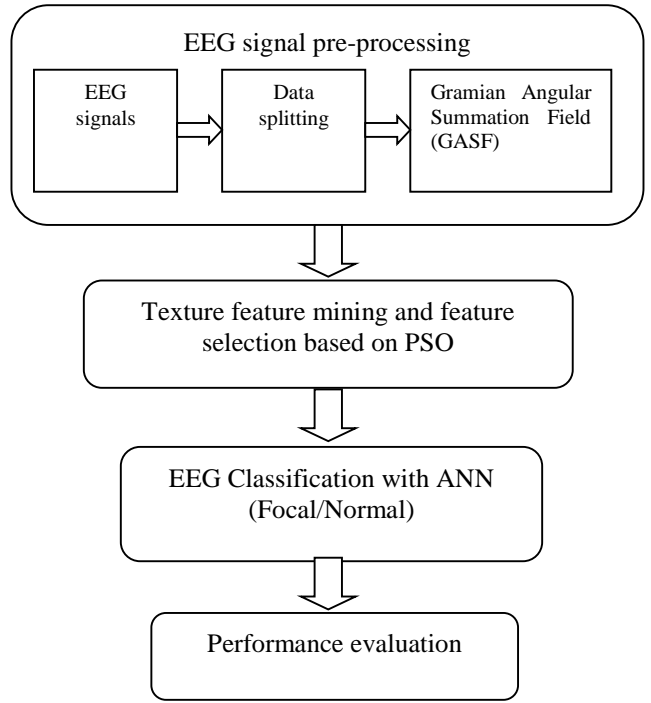

**Fig.4. Block diagram of Feature-Based Deep ANN for EEG assessment**

As shown in Fig. 4, this method consists of feature extraction, Particle Swarm Optimization (PSO) based dominant feature selection, and classification based on a customized ANN model. The mathematical expressions of these transforms can be expressed as follows;

- *ST*:

Consider a continuous-time sequence $H(t)$, with a spectrum sample $t - \tau$ can be found by $H(t) \cdot g(t-\tau, \sigma)$ where $g(t-\tau, \sigma) = \frac{1}{\sqrt{2\Pi\sigma}} e^{\frac{t^2}{2\sigma^2}}$ is the Gaussian-window at $\tau$.

Then, the ST in the frequency domain can be expressed as,

$$S(f, \tau, \sigma) = \int_{-\infty}^{\infty} H(t) g(t-\tau, \sigma) e^{-i2\Pi ft} dt \quad (4)$$

where, $\sigma = \frac{1}{f}$ is the dilation parameter of frequency and $e^{-i2\Pi ft}$ is the exponential kernel function. Additional details on ST can be found in [29].

- *WVT:*

Usually, the WVT is a double Fourier transform of regular uncertainty function. Mathematically, it can be expressed as;

$$P_x(t,f) = \int_{-\infty}^{\infty} e^{-i2\Pi f\tau} S^*(t - \frac{1}{2}\tau) \cdot S(t + \frac{1}{2}\tau) d\tau \quad (5)$$

where, $S^*(t)$ and $S(t)$ are the original and invented signals. Other details can be accessed from [28,29].

- *SET*:

SET is a recently proposed technique to examine the instantaneous magnitude and frequency of a test signal.

Let the signal component is expressed as;

$$S(t) = \sum_{k=1}^{t} A_k(t).e^{i\phi_k}(t) \quad (6)$$

Each of the signals is alienated based on distance and the window function, and it can be expressed as;

$$\phi_{k+1}(t) - \phi(t) > 2\Delta \quad (7)$$

where, $A_k$ =instantaneous magnitude of $k^{th}$ signal, $\varphi_k$ =instantaneous phase of the $k^{th}$ signal, and $\Delta$ =frequency of the window function. The details regarding SET is clearly described in [29].

- *STFT*:

This function is mathematically expressed as;

$$X(k) = \frac{1}{T} \sum_{t=0}^{T-1} w(t).x(t) e^{\frac{-i2\Pi}{T}kt} \quad (8)$$

where, $X(k)$ =frequency spectrum $k^{th}$ component, w(t)= window function, x(t)= signal sample till $t^{th}$ times, and T=sample number in window. Other information can be found in [29].

After extracting the relevant features from the GASF images of the EEG signals, a feature selection based on the PSO is implemented, as discussed in [29]. Then ten of dominant features, such as (i) ST: cluster-shade, sum-entropy, (ii) WVT: contrast, fractal-dimension, difference-variance, (iii) SET: auto-correlation, sum-average, and (iv) STFT: cluster-prominence, Information Measure of correlation, and homogeneity are selected. And these features are used to train and validate the ANN classifier.

**5.4 Performance Measures**

The performance of the various DNN methods discussed here is assessed by computing the performance metrics, as discussed in the literature [43].

The expressions for the performance measures are given below:

$$Precision = T_{+ve}/(T_{+ve} + F_{+ve}) \quad (9)$$

$$Recall = Sensitvity = T_{+ve}/(T_{+ve} + F_{-ve}) \quad (10)$$

$$F_1 - score = 2\left(\frac{Precision \times Recall}{Precision + Recall}\right) \quad (11)$$

where, $T_{-ve}$, $T_{+ve}$, $F_{-ve}$, and $F_{+ve}$ indicates true-negative, true-positive, false-negative, and false-positive, respectively.

When these values are close to unity or the percentage value is closer to 100, then the implemented technique is the best possible approach to examine the considered dataset.

## 6 Results and Discussion

In this section, we exhibit the experimental results attained from the adopted Deep-Learning (DL) approaches. Initially, the EEG signals are transformed into RGB scale images using the GASF technique. Later, the converted image is then assessed using the three DNN models, namely the pre-trained networks, Custom CNN, and the feature-based Deep ANN.

Initially, the two sample EEG records are considered, and the obtained result is depicted in Fig 5. It shows the Normal and Focal EEG time series data. Applying the GASF procedure to this time-series data, we obtain GASF images, which are shown in Fig. 6. From this figure, it can be noted that the spatial pattern of the GASF images of normal and focal EEG epoch shows subtle changes when the time duration of 2000 sample points is taken. Thus a data-splitting procedure is implemented to get the EEG image with a dimension 256x256 pixels, which are shown in Fig.7. This image shows detailed spatial features of the GASF for the normal and focal EEG signals. Thus, the EEG records are split into 256 samples for obtaining the GASF images for the normal and focal cases.

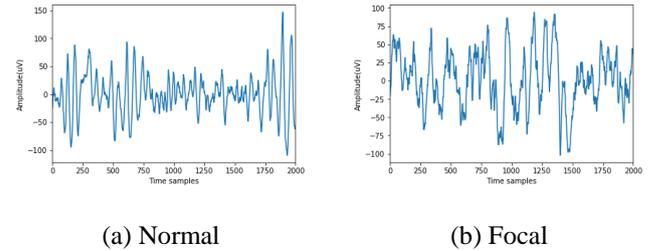

(a) Normal  (b) Focal

**Fig.5. Single channel EEG signal of (a) Normal and (b)Focal waves**

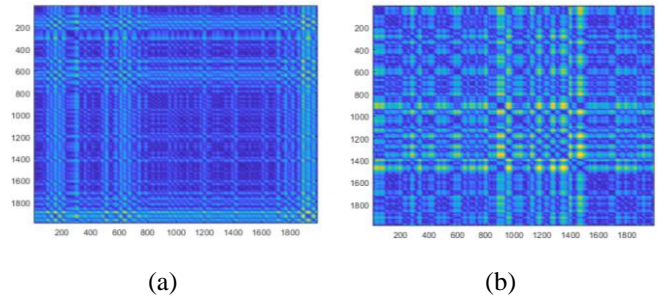

(a)  (b)

**Fig.6. Conversion of the EEG signal into the image (a) Normal, (b) Focal GASF**

Once the images were obtained using GASF, different DNN subroutine was run to validate the models for epilepsy detection. The AlexNet and Feature-based Deep ANN is run in the Matlab environment, and VGG and Custom CNN are developed in Python environment using open-source deep learning frameworks such as TensorFlow and Keras.

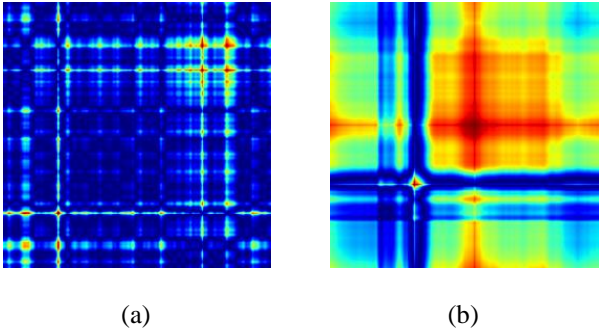

**Fig.7. Representation of Normal and Focal EEG signal using GASF after data splitting (a) Normal GASF (b) Focal GASF**

Performance metrics such as Precision, Recall, and F1-score are calculated to compare the considered DNN models for epilepsy detection. Table 2 summarizes the performance of the models for the normal and focal GASF detection. Moreover, an average of the performance metrics considering the normal and focal GASF is also provided. It can be noted that AlexNet provided an overall Precision of 0.7451, Recall of 0.7435, and F1-score of 0.7436. The performance is low compared to the other DNN models considered here. VGG16 and VGG19 models provided a similar classification performance for epilepsy detection from GASF images. They were reported with an F1-score of 0.7518 and 0.74995, respectively.

**Table 2 Quantitative performance analysis of various DNN networks in the classification of EEG based on GASF images**

| DL architecture | Signal | Precision | Recall | F1-score | AUC |
|---|---|---|---|---|---|
| AlexNet | Focal | 0.7314 | 0.772 | 0.7511 | 0.71 |
| | Normal | 0.758 6 | 0.715 | 0.7361 | |
| | **Average** | **0.7451** | **0.7435** | **0.7436** | |
| VGG16 | Focal | 0.7184 | 0.8655 | 0.7782 | 0.750 |
| | Normal | 0.8218 | 0.6493 | 0.7254 | |
| | **Average** | **0.7701** | **0.7574** | **0.7518** | |
| VGG19 | Focal | 0.8439 | 0.6218 | 0.7160 | 0.750 |
| | Normal | 0.7083 | 0.8851 | 0.7839 | |
| | **Average** | **0.7761** | **0.7534** | **0.74995** | |
| Feature based ANN | Focal | 0.8426 | 0.8046 | 0.7947 | 0.85 |
| | Normal | 0.8314 | 0.7996 | 0.8015 | |
| | **Average** | **0.8370** | **0.8021** | **0.7981** | |
| Custom CNN | Focal | **0.80** | **0.93** | **0.86** | **0.92** |
| | Normal | **0.97** | **0.91** | **0.94** | |
| | **Average** | **0.885** | **0.92** | **0.90** | |

Feature-Based DNN approach provided the best classification performance compared to other pre-trained models. It is reported here with an average Precision of 0.8370, Recall of 0.8021, and F1-score of 0.7981. However, we report here that the custom CNN based deep learning approach provided the highest classification performance compared to all other models considered here. It is reported with the highest Precision of 0.885, Recall of 0.92, and F1-score of 0.90. Moreover, the Receiver Operating Characteristic (ROC) curve is also plotted for all the models for the performance assessment using an alternative metric of Area under the Curve (AUC). For an ideal classifier, the AUC value is '1' and with a random guess taking an AUC value of 0.5 (given as dotted line on the ROC plots shown in Fig.8). All the DL models reported an AUC value of >0.5, which is better than a random guess. However, Custom based CNN model provided the highest AUC value of 0.92, which is better than other DNN models compared here.

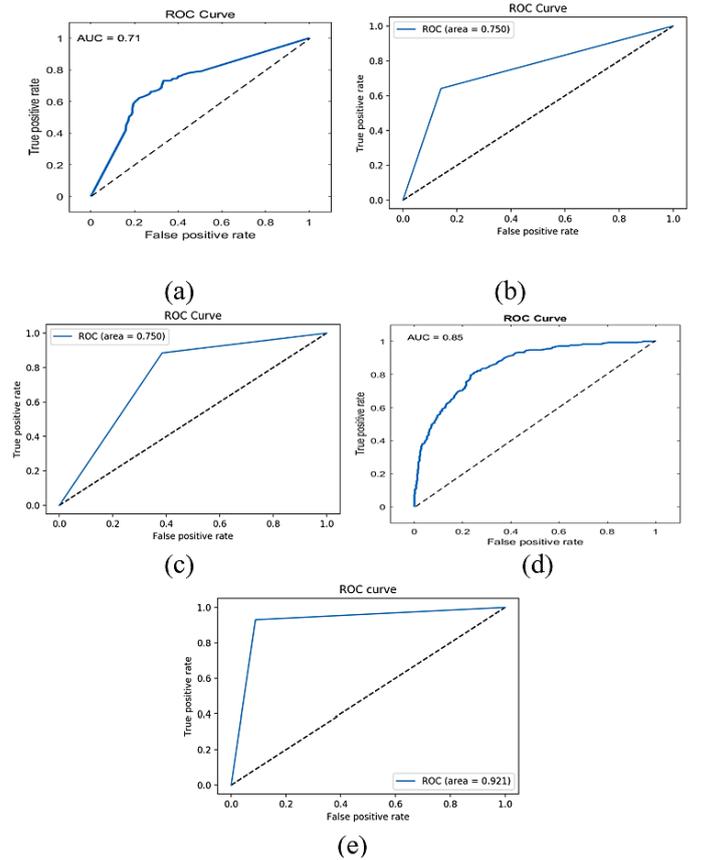

**Fig.8. ROC Curve of various DNN models considered here (a) AlexNet, (b) VGG16, (c) VGG19, (d) Feature-based ANN, (e) Custom CNN**

The epilepsy detection accuracy of the DNN models considered here can be compared to some of the epilepsy detection works done using the same Bern EEG dataset. Sharma et al. reported an accuracy of 87% based on feature extraction using Empirical Mode Decomposition (EMD) [20]. Das et al. proposed a combination of features from EMD and Discrete Wavelet Transform (DWT) and reported an accuracy of 89.04% [21]. Bhattacharyya et al. proposed an Empirical Wavelet Transform (EWT) for the classification of Focal EEG signals. They reported an overall accuracy of 90% [22]. It can be observed that custom CNN provides a better classification of Focal EEG signals from GASF images with a peak F1-score of 0.90 and an AUC value of 0.92, respectively.

The DNN models implemented in this work involves epilepsy diagnosis from imaging the EEG time series data. GASF images involve transforming the EEG time samples of length 256 and thus leading to a GASF image dimension of 256×256. This approach was tested with only the Bern EEG dataset. The ability of GASF to capture the temporal dependence of epilepsy in EEG is not fully experimented with different sample windows. Further investigations are required in testing the GASF approach with different clinical data for epilepsy diagnosis using deep learning approaches.

## 7 Conclusion

This work assessed the approach of transforming the EEG time series data to images using Gramian Angular Summation Field (GASF) for epilepsy detection. The work has exploited the advantages of using deep learning approaches in image classification tasks for EEG signal classification. Unlike conventional approaches of performing feature extraction and feature selection for epilepsy detection from EEG, this work used GASF images for detecting the focal episodes. We have used different DNN approaches such as pre-trained CNN models such as AlexNet, VGG16, and VGG19 for epilepsy Detection. Moreover, a feature-based ANN and a Custom CNN model are also implemented for the detection of focal signals through GASF images. This work reports that Custom CNN architecture with three convolution blocks with batch normalization and max-pooling layers have better classification performance in terms of Precision, Recall, and F1-score. This work provides an alternate approach of imaging EEG time series data for epilepsy detection. Here there is no manual intervention required for improving the classification accuracy as the feature extraction and feature selection stages are automated in the Custom CNN approach proposed here.

**Conflict of interest**

The authors declare that they have no conflict of interest